# Development of Direct Kinematics and Workspace Representation for Smokie Robot Manipulator & the Barret WAM

Reza Yazdanpanah A.

*Abstract*— **This paper discusses modelling two 6 DOF arm robots. The first step of modelling a robot is establishing its Denavit-Hartenberg parameters. It requires assigning proper coordinates for each link and finding their exact dimensions. In this project we will develop the direct kinematics and workspace representations for two manipulators: the Smokie Robot and the Barrett WAM. After finding the D-H parameters and creating Transformation Matrices,MATLAB programming is used to represent their workspaces.**

## I. Introduction

THIS paper discusses the direct kinematics and modelling of two serial manipulators, each with 6 degrees of freedom: the Smokie Robot and the Barrett Whole Arm Manipulator(WAM). A manipulator consists of a series of rigid bodies (links) connected by means of kinematic joints. The first step of modelling a robot is establishing its Denavit-Hartenberg parameters..In order to derive the D-H parameters we identify the joints and degrees of freedom of each robot. For assigning proper coordinates for each link, we follow Denavit-Hartenberg convention rules [1] by which a systematic, general method is used to define the relative position and orientation of two consecutive links.

In this paper first we shedlight on Smokie robot and WAM arm and choose the joints and links. Then the D-H parameters are derived and joint limits are specified. Next the workspace plots are shown for different views and finally the MATLAB code is explained.

## II. Manipulators

Manipulators are robots with a mechanical arm operating under computer control. They are composed of links that are connected by joints to form a kinematic chain. The manipulators that is considered in this thesis have solely rotary, also called revolute, joints. Each represents the interconnection between two links. The axis of rotation of a revolute joint, denoted by $z_i$, is the interconnection of links $l_i$ and $l_{i+1}$. The joint variables, denoted by $q_i$, represent the relative displacement between adjacent links

In this paper we study two Manipulators: the Smokie Robot and the Barrett Whole Arm Manipulation (WAM) arm. These two are introduced in this section.

### A. Smokie Robot OUR

The OUR is a low-cost, 6-DOF industrial manipulator manufactured by Smokie Robots. With a weight of 18.4 kg it is a lightweight manipulator. It has a reach of 85 cm and a maximal payload of 5 kg and is shown in Figure 1.

OUR is a very low cost robot that has the comparable performance with any general industrial robots. OUR's modularized design enables users to reconfigure the robot system with 4-7 DoFs to meet their requirements. The standard OUR is designed with 6-DoFs.

Open Unit Robot opens the development and integration of control systems to users. The users have easy access to controlling the robot through any equipment with real-time CAN-Bus.

### B. WAM Arm

The WAM Arm is a 7-degree-of-freedom (7-DOF) manipulator with human- like kinematics. With its aluminum frame and advanced cable-drive systems, including a patented cabled differential, the WAM is lightweight with no backlash, extremely low friction, and stiff transmissions. All of these characteristics contribute to its highbandwidth performance. The WAM Arm is the ideal platform for implementing Whole Arm Manipulation (WAM), advanced force control techniques, and high precision trajectory control. WAM is shown in figure 2.

The WAM Arm is a highly dexterous backdrivable manipulator. It is the only commercially available robotic arm with direct-drive capability supported by Transparent Dynamic between the motors and joints, so its joint-torque control is unmatched and guaranteed stable.

Reza Yazdanpanah Abdolmalaki is with The University of Tennessee Knoxville, TN 37996 USA (e-mail: ryazdanp@vols.utk.edu).

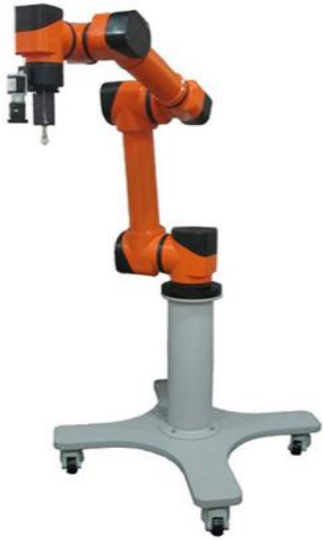

*Figure 2: OUR Smokie Robot*

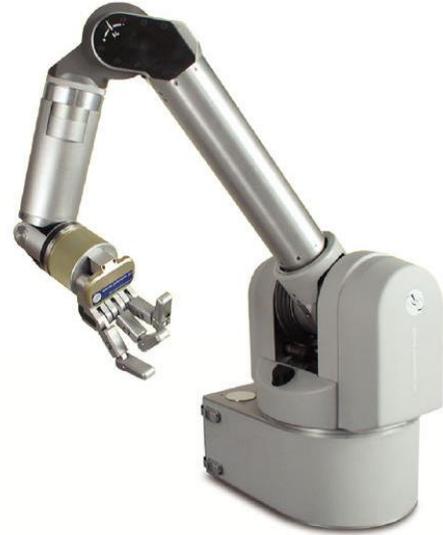

*Figure 1: WAM Arm*

## III. LINKS AND JOINTS IDENTIFICATION

In this section we identify the links and joints of each Arm and designate the degrees of freedom of both manipulators.

### A. OUR

OUR Smokie robot consists of 6 links and 6 revolute joints. Each joint connects two consecutive links to each other. All the 6 revolute joints have $-180°{\sim}180°$ rotation capability. The links, Joints and Dimensions are shown in figure 3.

### B. WAM Arm

WAM Arm robot has 7 DoFs. It has 7 links and 7 joints. In this study, by assuming the lower arm rotational joint as fixed we consider a 6DOF version of the WAM. The Joints and dimensions are shown in Figure 4. Table 1 demonstrates the joint rotation limits.
Although this Robot has 7 degrees of freedom, in this study we plan to study a 6 DoF version of this Arm, by assuming the lower arm rotational joint is fixed.

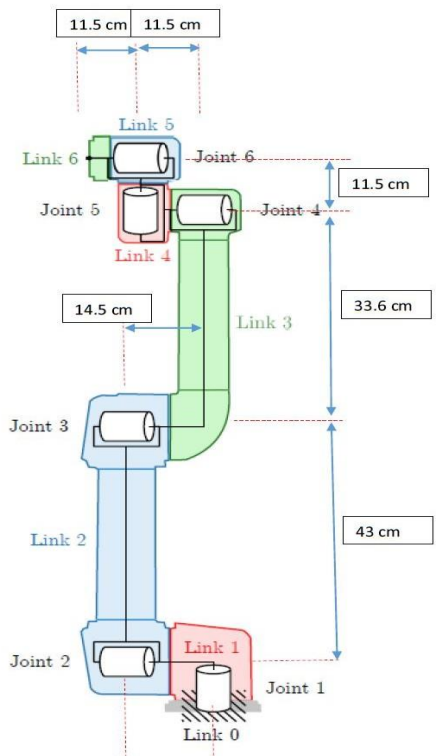

*Figure 4: OUR Arm Dimension, Joints and Links*

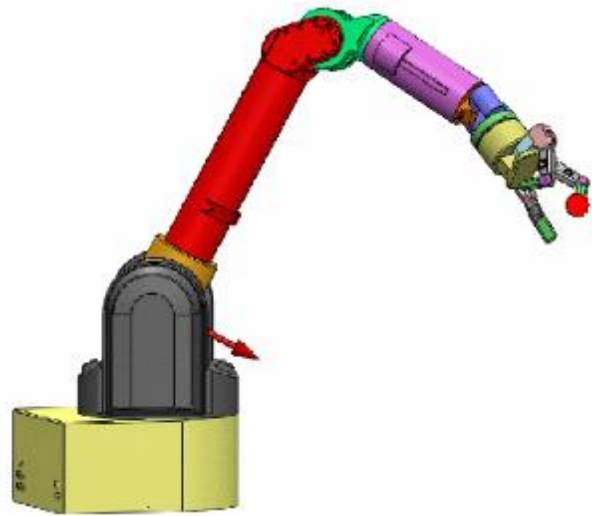

*Figure 3:4: WAM Arm links and joints*

## IV. Workspace Design of Smokie arm

### A. Denavit-Hartenberg Parameters

The commonly used DH-convention defines four parameters that describe how the reference frame of each link is attached to the robot manipulator. Starting with the inertial reference frame, one additional reference frame is assigned for every link of the manipulator. The four parameters $a_i, d_i, \alpha_i, \theta_i$ defined for each link $i \in [1, n]$ transforms reference frame $i-1$ to $i$ using the four basic transformations.

For this purpose we choose the coordinates as follows. The algorithm presented in D-H convention is used to assign the proper coordinates.

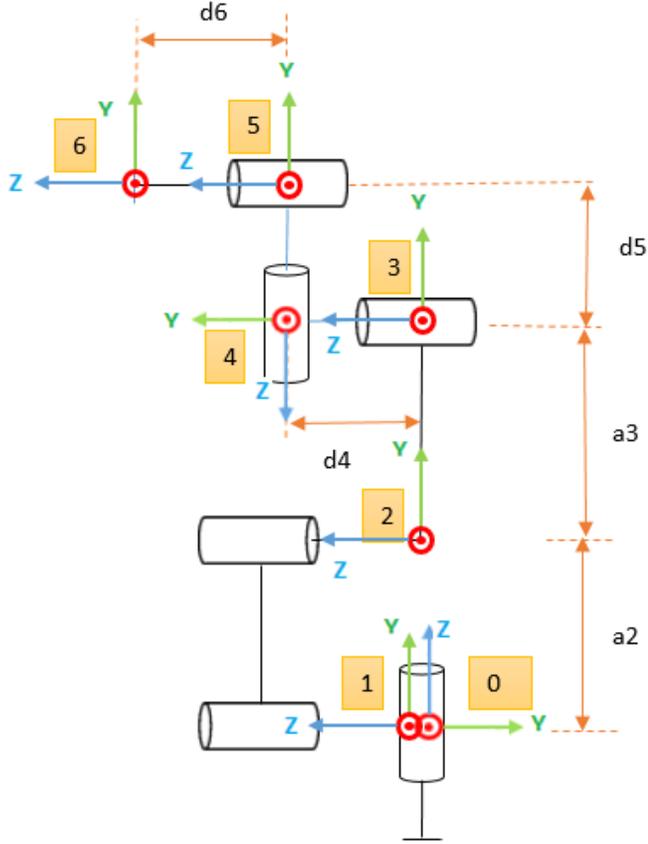

*Figure 5: D-H coordinates and dimensions*

After selecting the coordinates, the D-H parameters can be derived easily. The dimensions are measured by caliper and ruler.

*Table 1: D-H parameters of OUR Smokie arm*

| i | $a_i$ | $d_i$ | $\alpha_i$ | $\theta_i$ |
|---|---|---|---|---|
| 1 | 0 | 0 | $\pi/2$ | $\theta_1$ |
| 2 | a2= 43 | 0 | 0 | $\theta_2$ |
| 3 | a3= 33.6 | 0 | 0 | $\theta_3$ |
| 4 | 0 | d4= 11.5 | $\pi/2$ | $\theta_4$ |
| 5 | 0 | d5= 14.5 | $-\pi/2$ | $\theta_5$ |
| 6 | 0 | d6= 11.5 | 0 | $\theta_6$ |

### B. Transfer function of End Effector wrt Base

The transformation matrix from Frame i to Frame i-1 is a function only of the joint variable $q_i$, that is, $\theta_i$ for a revolute joint or $d_i$ for a prismatic joint. The Denavit-Hartenberg convention allows constructing the direct kinematics function by composition of the individual coordinate transformations into one homogeneous transformation matrix. The procedure can be applied to any open kinematic chain and can be easily rewritten in a programming code. [2]

The resulting coordinate transformation is:

$$A_i^{i-1}(q_i) = \begin{bmatrix} C\theta_i & -S\theta_i C\alpha_i & S\theta_i S\alpha_i & a_i C\theta_i \\ S\theta_i & C\theta_i C\alpha_i & -C\theta_i S\alpha_i & a_i S\theta_i \\ 0 & -S\alpha_i & C\alpha_i & d_i \\ 0 & 0 & 0 & 1 \end{bmatrix}$$

By running the MATLAB code (written for calculating the Transfer function and Plotting the workspace), the following:

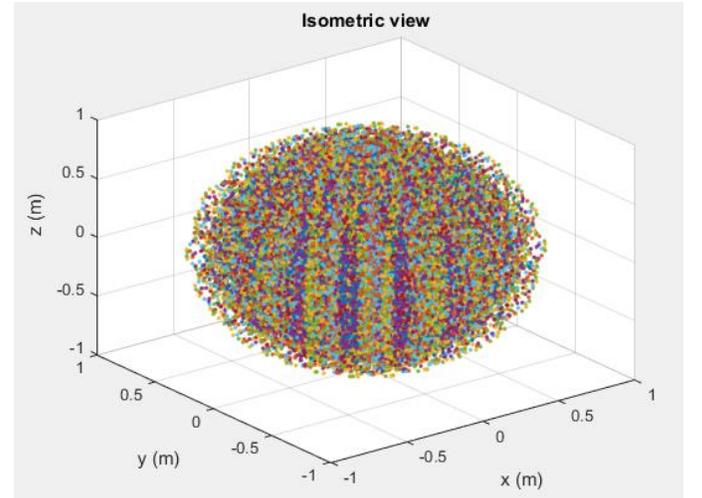

*Figure 6: Isometric view of Smokie workspace*

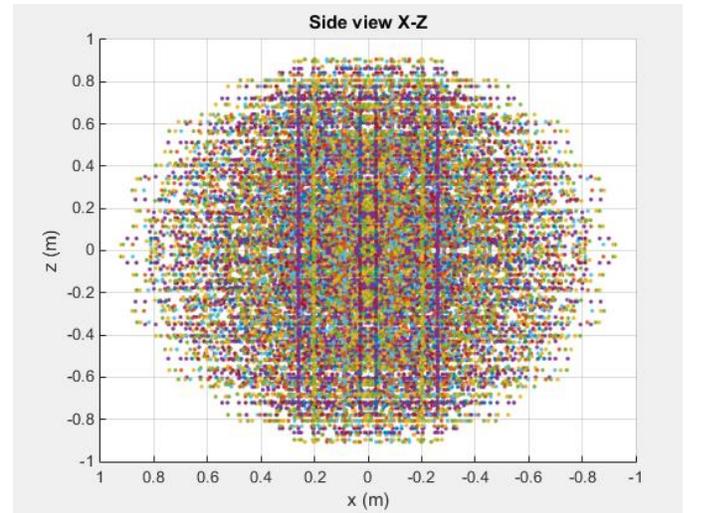

*Figure 7: Side view of Smokie workspace (X-Z)*

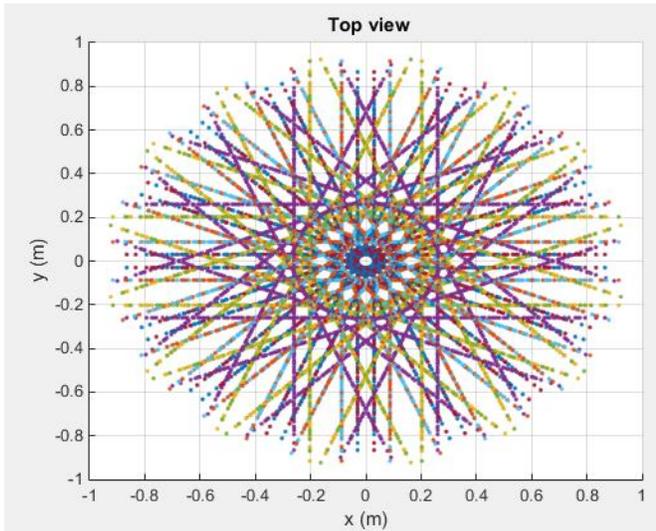

*Figure 8: Top-view of Smokie workspace (X-Y)*

## V. WORKSPACE DESIGN OF WAM

### A. Denavit-Hartenberg Parameters

Exactly the same as previous procedure, we will first assign the proper coordinates for WAM based on D-H convention.

WAM has 7 DoF, by assuming the first joint fixed, WAM would have 6 DoF and 7 coordinates should vbe chosen (i=0:6).

By assigning the coordinates, The D-H parameters can be calculated and are in the Table 2.

The first line of Table 2 is deleted for this study because of fixed joint.

*Table 2: D-H parameters WAM*

| i | $a_i$ | $d_i$ | $\alpha_i$ | $\theta_i$ |
|---|---|---|---|---|
| 1 | 0 | 0 | $-\pi/2$ | $\theta_1$ |
| 2 | 0 | 0 | $\pi/2$ | $\theta_2$ |
| 3 | a3= 0.045 | d3=0.55 | $-\pi/2$ | $\theta_3$ |
| 4 | a4=-0.045 | 0 | $\pi/2$ | $\theta_4$ |
| 5 | 0 | d5= 0.3 | $-\pi/2$ | $\theta_5$ |
| 6 | 0 | 0 | $\pi/2$ | $\theta_6$ |
| 7 | 0 | d6=0.06 | 0 | $\theta_7$ |

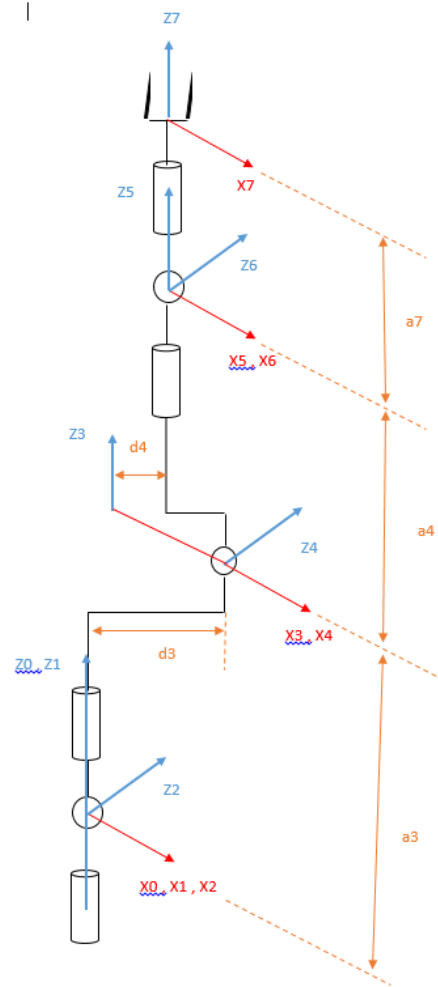

*Figure 9: WAM Coordinates and dimensions*

### B. Finding Joint Limits

The joints of WAM have angle limits. For assigning the range of angle of each joint, The joint limits are come in Table3.

*Table 3: Joint limits of WAM*

| joint | Min Angle Limit (Rad) | Max. Angle limit (Rad) |
|---|---|---|
| 1 | -2.6 | 2.6 |
| 2 | -2.0 | 2.0 |
| 3 | -2.8 | 2.8 |
| 4 | -0.9 | 3.1 |
| 5 | -4.8 | 1.3 |
| 6 | -1.6 | 1.6 |
| 7 | -2.2 | 2.2 |

## C. Workspace Plots

By using the coordinate transformation matrix and coding in MATLAB , the workspace of 6 DoF is plotted.
Different views of the workspace is plotted in below figures.

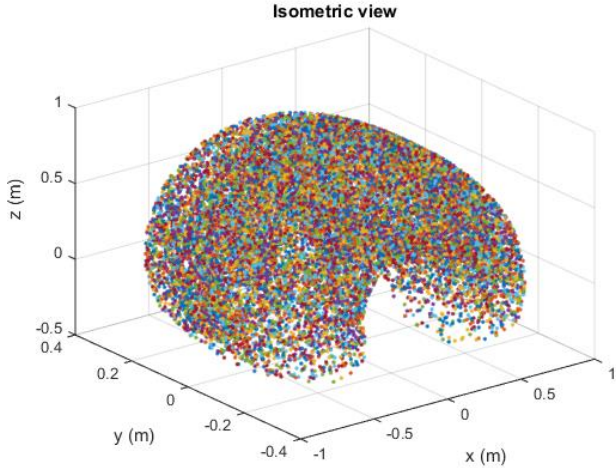

Figure 10: Isometric view of WAM's Workspace

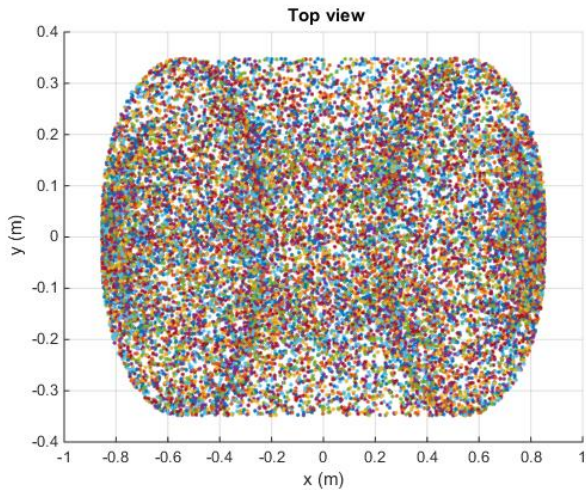

Figure 11: Top view of WAM's Workspace

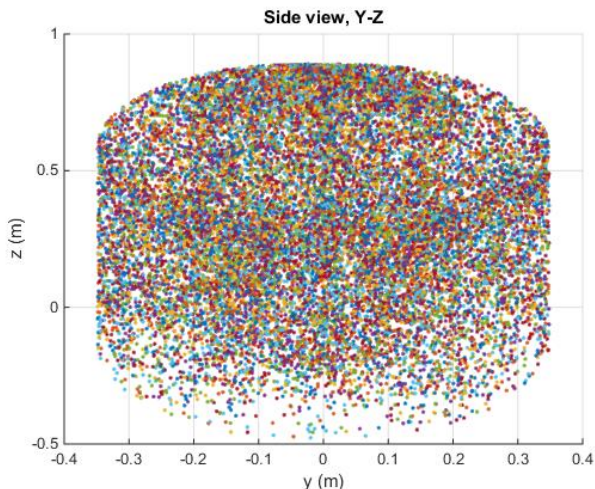

Figure 12: Side view of WAM's Workspace (Y-Z)

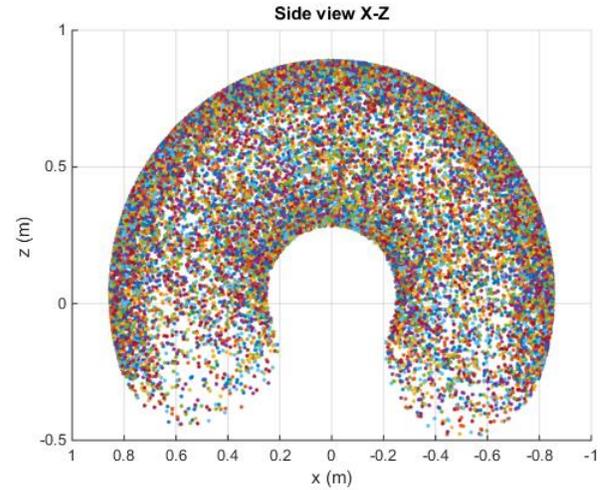

Figure 13: Side view of WAM's Workspace (X-Z)

## VI. MATLAB CODE EXPLANATION

### A. Inserting D-H Parameters

In this part we insert the D-H parameters manually to the code.

```
% Inserting D-H convention parameters
a1 = 0;      alpha1 = pi/2;   d1 = 0.0345;
a2 = 0;      alpha2 = -pi/2;  d2 = 0;
a3 =-0.045;  alpha3 = pi/2;   d3 = 0.55;
a4 = 0.045;  alpha4 = -pi/2;  d4= 0;
a5 = 0;      alpha5 = -pi/2;  d5 = 0.30;
a6 = 0;      alpha6 = pi/2;   d6 = 0;
a7 = 0;      alpha7 = 0;      d7 =0.060;
```

### B. Inserting the joint limits

In this part we insert the angle limits of each joint for each Arm.

```
% Inserting joint limits for Arms
t1=0
t2_min = -1.9;   t2_max = 1.9;
t3_min = -2.8;   t3_max = 2.8;
t4_min = -0.9;   t4_max = 3.14;
t5_min = -4.8;   t5_max = 1.3;
t6_min = -1.6;   t6_max = 1.6;
t7_min = -2.2;   t7_max = 2.2;
```

### C. Monte Carlo Method for plotting

For plotting the workspace, we use the Monte-Carlo method. This method use random intervals between the minimum Angle and Maximum Angle. We can define the number of samples in each interval. We choose N for the number of intervals.

```
% Monte Carlo method
% sampling size
N = 20000;
t1 = t1_min + (t1_max-t1_min)*rand(N,1);
t2 = t2_min + (t2_max-t2_min)*rand(N,1);
t3 = t3_min + (t3_max-t3_min)*rand(N,1);
t4 = t4_min + (t4_max-t4_min)*rand(N,1);
t5 = t5_min + (t5_max-t5_min)*rand(N,1);
t6 = t6_min + (t6_max-t6_min)*rand(N,1);
t7 = t7_min + (t7_max-t7_min)*rand(N,1);
```

*D. Transformation Matrix creation*

In order to create transformation matrix we use a function named TransMat.

```
function [ T ] = TranMat( a,b,c,d )

T = [ cos(d)   -sin(d)*cos(b)
sin(d)*sin(b)  a*cos(d); sin(d)
cos(d)*cos(b)  -cos(d)*sin(b)   a*sin(d);
0 sin(b)   cos(b)    c;
0   0    0   1
    ];
end
```

*E. Plotting the workspace*

For plotting the workspace, we first create each transfer Matrix $A_i^{i-1}$ and by postmultiplying them, we have our Transformation matrix of $A_{ee}^0$ named as T.
Now the program Plots th fourth column of matrix T for all our samples (N). this plot done in the same loop as before.
We use plot 3 for this purpose and each step, our program gets 3 magnitudes of X,Y,Z and plots them in figure.

```
for i = 1:N
 A1 = TransMat(a1,alpha1,d1,t1(i));
 A2 = TransMat(a2,alpha2,d2,t2(i));
 A3 = TransMat(a3,alpha3,d3,t3(i));
 A4 = TransMat(a4,alpha4,d4,t4(i));
 A5 = TransMat(a5,alpha5,d5,t5(i));
 A6 = TransMat(a6,alpha6,d6,t6(i));
 A7 = TransMat(a7,alpha7,d7,t7(i));
 T = A1*A2*A3*A4*A5;
 X=T(1,4);
 Y=T(2,4);
 Z=T(3,4);
 plot3(X,Y,Z,'.')
 hold on;
end
```

This plot3 gives us the isometric view of our workspace.
For having top-view and side-views, we use the command: view in MATLAB. By labeling the axes and choosing title, we have our final plots.

```
view(3);
title('Isometric view');
xlabel('x (m)');
ylabel('y (m)');
zlabel('z (m) ');

view(2); % top view
title(' Top view');
xlabel('x (m)');
ylabel('y (m)');

view([1 0 0]); % y-z plane
title('Side view, Y-Z');
ylabel('y (m)');
zlabel('z (m)');
```

By increasing N we have figures with better resolution. But the Time of running increases also.
It's preferred to use View command, because handling the figures is a heavy process when we have great N.

## VII. CONCLUSION AND DISCUSSION

In this Project paper we used computer programming for finding the Workspace Volume of two arm robots. Both WM and Smokie arms are 6 DoFs in this study. The overall algorithm for this purpose is :
:

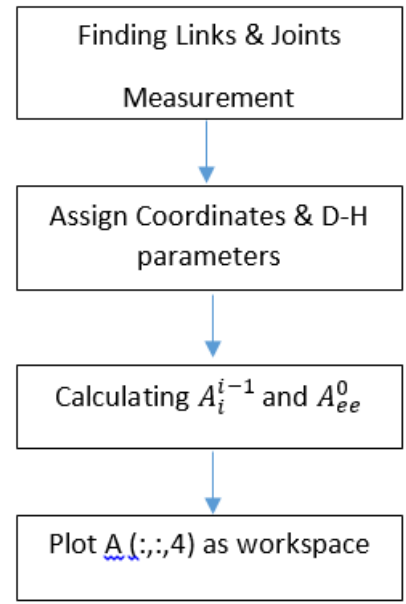

Since smokie robot is a new version, there is no source to compare. Even the company didn't publish the workspace figure in their Specification sheet and website.
For WAM I run the program for & DoF version to have a resource.By comparing the results for 7 DoF with the workspace which company published, a very good similarity was found.
For plotting the workspace Monte Carlo Method is used, this method is more efficient, since it uses the random angles in the joint angle range and we can define the number of samples.